\documentclass[10pt]{article}
\usepackage[preprint]{rlj} 
\usepackage{amssymb}            
\usepackage{mathtools}          
\usepackage{mathrsfs}           
\usepackage{graphicx}           
\usepackage{subcaption}         
\usepackage[space]{grffile}     
\usepackage{url}                
\usepackage{lipsum}     
\usepackage{booktabs}

\title{Quantifying First‐Order Markov Violations in Noisy Reinforcement Learning: A Causal Discovery Approach}
\setrunningtitle{Markovianess}
\author{Naveen Mysore}
\emails{nmysore@ucsb.edu}
\affiliations{
\textbf{Department of Electrical and Computer Engineering, University of California, Santa Barbara, USA}
}

\contribution{
    A new normalized Markov Violation Score (MVS) to quantify the degree of non-Markovian dependence at lags beyond one in high-dimensional environments under observational noise
}{
    Although prior methods acknowledge the effects of noise or partial observability, they often lack a single scalar measure that captures multi-lag dependencies. The proposed MVS addresses this gap by providing a unified, quantifiable metric of Markov violations.
}

\contribution{
    Empirically demonstrates how MVS correlates with the performance of an RL agent (PPO) across varying noise levels, thereby linking Markovian-ness to learning outcomes
}{
    Prior works have suggested that noisy or partial observability can degrade RL performance, but explicit correlations with quantitative Markovian-ness metrics remain underexplored.
}

\contribution{
    Shows that dimension manipulations derived from causal discovery (e.g., removing low-influence dimensions) minimally degrade performance, while selectively adding extra noise to highly influential dimensions significantly lowers rewards
}{
    While dimensionality reduction is often done heuristically, this approach provides causal evidence for which dimensions remain critical under noise, illustrating a targeted approach to environment design and state abstraction.
}
\summary{
This work examines how observational noise degrades the Markov property in reinforcement learning (RL) environments and measures its impact on policy performance. By applying the PCMCI causal discovery technique to detect multi‐lag dependencies, a novel Markov Violation Score (MVS) is introduced to quantify deviations from first‐order Markov assumptions. Experiments on classic control tasks (e.g., CartPole, Pendulum) under various noise levels reveal that higher MVS values typically correlate with reduced PPO policy performance.

Dimension manipulation guided by causal insights shows that removing low‐influence state dimensions exerts minimal impact on returns, whereas injecting extra noise into “crucial” dimensions significantly lowers performance. These findings demonstrate how noise undermines Markovian structure and underscore the value of causal analysis for improving both interpretability and robustness in noisy RL settings.
}

\begin{document}
\maketitle 
\begin{abstract}
Reinforcement learning (RL) methods frequently assume that each new observation completely reflects the environment’s state, thereby guaranteeing Markovian (one‐step) transitions. In practice, partial observability or sensor/actuator noise often invalidates this assumption. The present paper proposes a systematic methodology for detecting such violations, combining a partial correlation‐based causal discovery process (PCMCI) with a novel Markov Violation score (MVS). The MVS measures multi‐step dependencies that emerge when noise or incomplete state information disrupts the Markov property.

Classic control tasks (CartPole, Pendulum, Acrobot) serve as examples to illustrate how targeted noise and dimension omissions affect both RL performance and measured Markov consistency. Surprisingly, even substantial observation noise sometimes fails to induce strong multi‐lag dependencies in certain domains (e.g., Acrobot). In contrast, dimension‐dropping investigations show that excluding some state variables (e.g., angular velocities in CartPole and Pendulum) significantly reduces returns and increases MVS, while removing other dimensions has minimal impact.

These findings emphasize the importance of locating and safeguarding the most causally essential dimensions in order to preserve effective single‐step learning. By integrating partial correlation tests with RL performance outcomes, the proposed approach precisely identifies when and where the Markov assumption is violated. This framework offers a principled mechanism for developing robust policies, informing representation learning, and addressing partial observability in real‐world RL scenarios. All code and experimental logs are accessible for reproducibility (\url{https://github.com/ucsb/markovianess}).
\end{abstract}
\section{Introduction}
\label{sec:introduction}

Reinforcement learning (RL) conventionally assumes that each observation completely encodes the environment’s state, thereby guaranteeing single‐step (Markov) transitions \citep{sutton_reinforcement_1998}. In practical scenarios, however, partial observability or sensor inaccuracies frequently violate this assumption \citep{wisniewski_benchmarking_2024}, leading to multi‐step correlations and weakened training outcomes. Although numerous RL methods can withstand mild noise, moderate or poorly structured perturbations tend to disrupt Markovian structure and erode policy performance.

A principal difficulty is in \emph{identifying} the precise circumstances under which the Markov property fails. Standard benchmarks (e.g., final returns) do not indicate whether an environment is effectively “non‐Markovian” from the learning agent’s standpoint. To address this, the current study introduces a \emph{Markov Violation Score} (MVS), leveraging partial correlation tests via PCMCI~\citep{runge_discovering_2022}. This metric highlights significant lag‐2‐and‐beyond correlations that point to multi‐step processes beyond first‐order assumptions.

A thorough examination explores how particular distortions affect both learning performance and MVS in three classic control tasks: \emph{CartPole-v1}, \emph{Pendulum-v1}, and \emph{Acrobot-v1}.
\begin{itemize}
    \item \textbf{Noise Injection.} Gaussian noise and autoregressive processes are injected into observation streams at various magnitudes, revealing which components are indispensable for stable operation.
    \item \textbf{Dimension Dropping.} Specific observation dimensions are removed entirely, compelling agents to learn in the presence of incomplete states. While some omissions lead to modest performance reductions, others trigger severe instability and elevated MVS.
    \item \textbf{Markov Violation Analysis.} For each manipulation, PCMCI detects higher‐order correlations (lags of 2 or more). Notable surges in multi‐step links typically align with abrupt performance drops, indicating that single‐step Markov assumptions no longer hold.
\end{itemize}

The outcomes show that different state components do not uniformly maintain Markovian structure. Altering or omitting a \emph{crucial} variable can lead to strong multi‐lag dependencies and dramatic policy breakdowns, while discarding less essential parts can have minimal impact. Moreover, each task exhibits distinct thresholds of resilience: certain environments degrade under moderate noise, whereas others (e.g., Acrobot) accommodate multi‐lag interactions without catastrophic failure.

\textbf{Paper Organization.}
Sections~\ref{sec:related-works}--\ref{sec:pcmci-markov} provide background on partial observability and causal discovery, then present the Markov property, PCMCI, and the introduced MVS. Section~\ref{sec:experiments-results} outlines the experimental procedure (baseline runs, noise perturbations, dimension dropping) and reports empirical findings on policy performance alongside Markov consistency. Section~\ref{sec:limitations-future} discusses constraints and outlines possible future directions, and Section~\ref{sec:conclusion} offers concluding remarks.
\section{Related Works}
\label{sec:related-works}

Real‐world reinforcement learning (RL) often faces partial observability and noisy signals that undercut the strict Markov property \citep{wiering_partially_2012}. In the domain of \emph{robust RL}, a variety of works address disturbances in observations or transitions~\citep{panaganti_robust_2022, liu_distributionally_2022}, employing adversarial training~\citep{pinto_robust_2017} or domain randomization~\citep{wang_robust_2019, li_mural_2021, wang_reinforcement_2020} to tackle noisy sensing. Some investigations incorporate noise directly into states or actions~\citep{hollenstein_colored_2024, hollenstein_action_nodate, igl_generalization_2019}, yet their evaluations frequently rely on final‐return measures and overlook a principled way to detect multi‐step dependencies arising from the breakdown of Markov assumptions.

Another area of \emph{partially observable RL} examines how unobserved factors degrade Markovian structure~\citep{lauri_partially_2023}. Within POMDPs and similar formalisms, latent variables \citep{liu_when_2022} often characterize environment dynamics~\citep{zhu_causal_2020, yu_two-way_nodate, shi_does_2020}. While such techniques can handle certain forms of noise (e.g., Gaussian or autoregressive), few methods pinpoint \emph{which} dimensions or temporal segments are crucial for sustaining or undermining first‐order dynamics. Moreover, a single, cohesive metric for quantifying multi‐lag correlations remains elusive.

In contrast, some studies \citep{ota_can_2020} expand input dimensionality to boost sample efficiency and final returns, highlighting the importance of retaining essential state information in larger feature spaces. However, these works do not clarify \emph{which} dimensions are strictly necessary for preserving a Markovian environment.

To address such gaps, the present study applies PCMCI’s causal discovery approach~\citep{runge_discovering_2022} to expose higher‐lag partial correlations and measure departures from the Markov property. Building on robust RL’s focus on noisy observations and partial‐observability research on hidden variables, this paper introduces a \emph{Markov Violation Score} (MVS) that consolidates multi‐step influences exceeding first‐order transitions. Unlike previous causal‐discovery research \citep{zeng_survey_2023} and partial‐observation studies, the MVS offers a single, interpretable metric reflecting how strongly the Markov assumption is broken under dimension removal or other perturbations. Consequently, this perspective surpasses reliance on final‐return metrics to identify precisely \emph{which} dropped variables or noise processes most severely compromise first‐order RL learning.
\section{Preliminaries}

\subsection{Markov Property and Markov Decision Processes}
\label{subsec:markov-mdp}

A discrete-time stochastic process \(\{X_t\}_{t=0}^{\infty}\) satisfies the \emph{Markov property}\ if, at every time step \(t\), the future state \(X_{t+1}\) is conditionally independent of all prior states \(\{X_0, X_1, \ldots, X_{t-1}\}\) given the current state \(X_t\). Formally,
\[
    P\bigl(X_{t+1} \mid X_t, X_{t-1}, \ldots, X_0\bigr)
    \;=\;
    P\bigl(X_{t+1} \mid X_t\bigr).
\]
Intuitively, this means the present state fully encapsulates all relevant information from the past. In a reinforcement learning (RL) context, we typically apply the Markov property to a state variable \(S_t\). If the environment truly satisfies this property, then
\[
    P\bigl(S_{t+1} = s', R_{t+1} = r \,\mid\, S_t = s, A_t = a, \ldots, S_0, A_0\bigr)
    \;=\;
    P\bigl(S_{t+1} = s', R_{t+1} = r \,\mid\, S_t = s, A_t = a\bigr),
\]
which ensures that only the current state \(S_t\) and action \(A_t\) determine the distribution over next states \(S_{t+1}\) and rewards \(R_{t+1}\). However, if noise or partial observability reduce the completeness of \(S_t\), higher-order (multi-lag) dependencies may arise. This violates the first-order Markov assumption and can complicate RL methods that rely on single-step dynamics.

\subsubsection{Conditional Independence and the PCMCI Framework}
\label{subsec:cond-indep-pcmci}

Two variables \(X\) and \(Y\) are said to be \emph{conditionally independent} given a set of variables \(Z\) if
\[
    P(X \,\mid\, Y, Z) \;=\; P(X \,\mid\, Z).
\]
In an ideal Markov process, once the current state \(S_t\) is known, the future state \(S_{t+1}\) becomes independent of all past states \(\{S_0, \dots, S_{t-1}\}\). However, noise or partial observability can introduce multi-lag dependencies, causing \(S_{t+1}\) to depend on earlier states \(S_{t-2}, S_{t-3}, \dots\). To detect such higher-order effects, one can examine \emph{partial correlations}, which measure linear associations between \(X\) and \(Y\) after conditioning on \(Z\). Significant partial correlations at lag~\(\ge 2\) indicate a breakdown of the first-order Markov property.

Constraint-based causal discovery methods, such as the \textbf{PC algorithm}~\citep{spirtes_causation_2001}, iteratively test for conditional independence and remove edges in a candidate causal graph. \emph{Momentary Conditional Independence (MCI)} extends this testing to time-series data by conditioning on momentary and past information at each time step. Building on MCI, \textbf{PCMCI}~\citep{runge_discovering_2022} combines partial-correlation-based tests with the PC procedure to handle high-dimensional time series. In an RL setting, detecting edges at lag~2 or beyond via PCMCI offers direct evidence that single-step conditioning on \(S_t\) alone is insufficient, thus revealing violations of the Markov property.

\paragraph{Relevance to RL and Markov Violations.}
In RL, $S_{t+1}$ often depends on $(S_t,A_t)$ only. Noise or partial observability can generate dependence on $S_{t-2}, S_{t-3}, \dots$ beyond $S_{t-1}$. By applying PCMCI to agent trajectories, one can quantify the severity of these multi-lag links. Such diagnosis helps explain policy breakdowns and suggests solutions like state augmentation or sensor fusion~\citep{laskin_reinforcement_2020}.

\subsection{PCMCI and the Markov Property}
\label{sec:pcmci-markov}

In a strictly Markovian environment, no significant causal links appear at lags beyond one. When PCMCI detects higher-lag correlations, it indicates missing information in $S_t$. After training, rollouts were collected to apply PCMCI across $S_{t-1}, S_{t-2},\dots$ to find significant partial correlations at $k \ge 2$. The \emph{Markov Violation Score} (Section~\ref{sec:markov-violation}) summarizes these multi-lag dependencies. Higher scores typically signal greater departure from first-order dynamics, aligning with observed performance drops.

\begin{table}[htbp]
    \centering
    \begin{tabular}{c c r r r}
    \toprule
    \textbf{Child} & \textbf{Parent} & \textbf{Lag} & \textbf{p-val} & \textbf{Part. Corr} \\
    \midrule
    \multicolumn{5}{l}{\emph{Variable 0 has 6 link(s):}}\\
    0 & 2 & 0   & 0.00000 & -0.833 \\
    0 & 3 & 0   & 0.00000 & -0.621 \\
    0 & 1 & 0   & 0.00000 &  0.566 \\
    0 & 0 & -1  & 0.00000 &  0.423 \\
    0 & 1 & -1  & 0.00000 &  0.109 \\
    0 & 2 & -1  & 0.00000 &  0.079 \\
    \bottomrule
    \end{tabular}
    \caption{An example of PCMCI results for CartPole showing no significant edges (p-value threshold was 0.05) at lag $\le -2$, consistent with first-order Markov structure in the unperturbed setting.}
    \label{tab:cartpole_pcmci_signif}
\end{table}
\section{Markov Violation Score}
\label{sec:markov-violation}

As noted in Section~\ref{sec:pcmci-markov}, PCMCI can reveal higher-lag dependencies that indicate violations of the first-order Markov property. This section introduces the \emph{Markov Violation Score} (MVS), which quantifies how severely one-step assumptions are broken.

\paragraph{Defining the MVS.}
Consider \(N\) total variables (e.g., state components), a maximum lag \(\tau_{\max}\), and a significance threshold \(\alpha_{\mathrm{level}}\). For each variable pair \((i,j)\) and lag \(|k| \ge 2\), let \(\textbf{val}_{(i,j,k)}\) be the partial correlation at lag~\(k\), and let \(\textbf{p}_{(i,j,k)}\) be its p-value. The indicator \(\mathbb{I}\bigl(\textbf{p}_{(i,j,k)} \le \alpha_{\mathrm{level}}\bigr)\) is 1 if the p-value is below \(\alpha_{\mathrm{level}}\) and 0 otherwise. The MVS then is
\[
  \mathrm{MVS}
    \;=\;
      \frac{
          \sum_{i=1}^{N} \sum_{j=1}^{N} \sum_{k=2}^{\tau_{\max}}
              (k-1)\,\bigl|\textbf{val}_{(i,j,k)}\bigr|\,
              \bigl[-\ln\bigl(\textbf{p}_{(i,j,k)}\bigr)\bigr]\,
              \mathbb{I}\bigl(\textbf{p}_{(i,j,k)} \le \alpha_{\mathrm{level}}\bigr)
      }
      {
          N^2\,\sum_{k=2}^{\tau_{\max}}(k-1)
      },
\]
where \((k-1)\) weights longer lags more heavily. If no lag\(|k|\ge 2\) links are detected, then \(\mathrm{MVS}=0\).

\begin{table}[htbp]
    \centering
    \footnotesize
    \begin{tabular}{lcccc}
        \toprule
        \textbf{Child Var} & \textbf{Parent Var} & \textbf{Lag} & \textbf{p-value} & \textbf{Partial Corr}\\
        \midrule
        \multicolumn{5}{l}{\emph{Variable 0 has 4 link(s):}}\\
         0 & 0 & -1 & 0.00000 &  0.663 \\
         0 & 3 & -3 & 0.00000 & -0.281 \\
         0 & 2 & -3 & 0.00000 & -0.078 \\
         0 & 1 &  0 & 0.03875 & -0.003 \\
        \bottomrule
    \end{tabular}
    \caption{Example PCMCI results ($\alpha$ threshold was 0.05) for a noisy CartPole run with \(\mathrm{MVS} > 0\).}
    \label{tab:mvs_nonzero_example}
\end{table}

\noindent
A nonzero MVS indicates multi-step dependencies that degrade performance in one-step RL algorithms. Larger scores correlate with more severe Markov violations, whereas \(\mathrm{MVS}=0\) means no multi-lag links survive thresholding and the system remains effectively first-order.
\section{Experiments and Results}\label{sec:experiments-results}
This section explores how noise injection and dimension manipulation impact both the Markovian structure of classic RL environments and final policy performance. The following subsections detail the experimental setup, baseline (no‐modification) runs, the effects of i.i.d.\ and autoregressive (AR) noise, and the consequences of dropping specific dimensions. Each analysis leverages both episode returns and the proposed Markov Violation Score (MVS) to reveal whether multi‐lag dependencies emerge under different perturbations.

\subsection{Experimental Setup}\label{sec:setup}
All experiments were implemented in Python~3 using \texttt{stable-baselines3}~\citep{ran_stable-baselines3_nodate} to train PPO~\citep{schulman_proximal_2017} agents with two‐hidden‐layer MLP policies. Each condition—be it a baseline run, noise injection, or dimension dropping—was initialized from scratch with a distinct random seed to ensure reproducibility. Training proceeded for 50k steps on \emph{CartPole-v1} and \emph{Acrobot-v1} (CartPole typically solves in 20k–30k but was extended to 50k for consistency), while \emph{Pendulum-v1} required 450k steps to approach near‐optimal swing‐up performance. Two primary metrics were recorded throughout: (1) \emph{episode returns} to measure policy performance, and (2) PCMCI outputs (partial correlations and p-values at lags \(k=1,\dots,5\)), used to compute the \emph{Markov Violation Score} (MVS). After each run, an additional 1000--2000 steps were sampled for PCMCI’s causal discovery, enabling direct comparisons of final returns and lag‐\(\ge2\) dependencies.

\subsection{Baseline Performance}\label{sec:baseline}
A no‐modification “baseline” was trained for each environment to verify that the default tasks exhibit effectively Markovian structure. In all three domains, the baseline (indicated by black curves in subsequent figures) converged quickly and maintained top returns, with PCMCI detecting negligible lag‐\(\ge2\) correlations (i.e., MVS~\(\approx0\)). This outcome confirms that the unaltered state representations of \emph{CartPole-v1}, \emph{Pendulum-v1}, and \emph{Acrobot-v1} largely satisfy first‐order Markov assumptions.

\subsection{Gaussian Noise Injection}\label{subsec:gaussian-noise-method}
\begin{figure}[htbp]
  \centering
  \begin{subfigure}[b]{0.45\textwidth}
    \includegraphics[width=\linewidth]{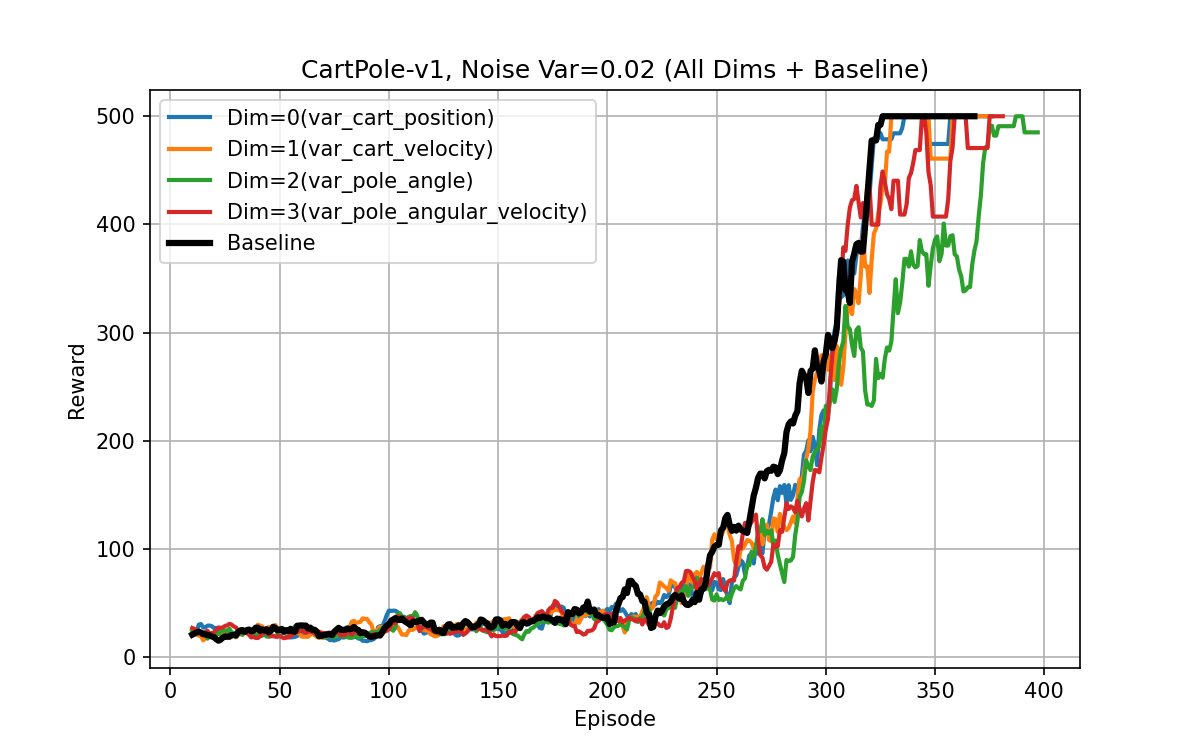}
    \caption{CartPole, $\mu=0$, $\sigma^2=0.02$}
    \label{fig:cartpole_noise_0.02}
  \end{subfigure}
  \begin{subfigure}[b]{0.45\textwidth}
    \includegraphics[width=\linewidth]{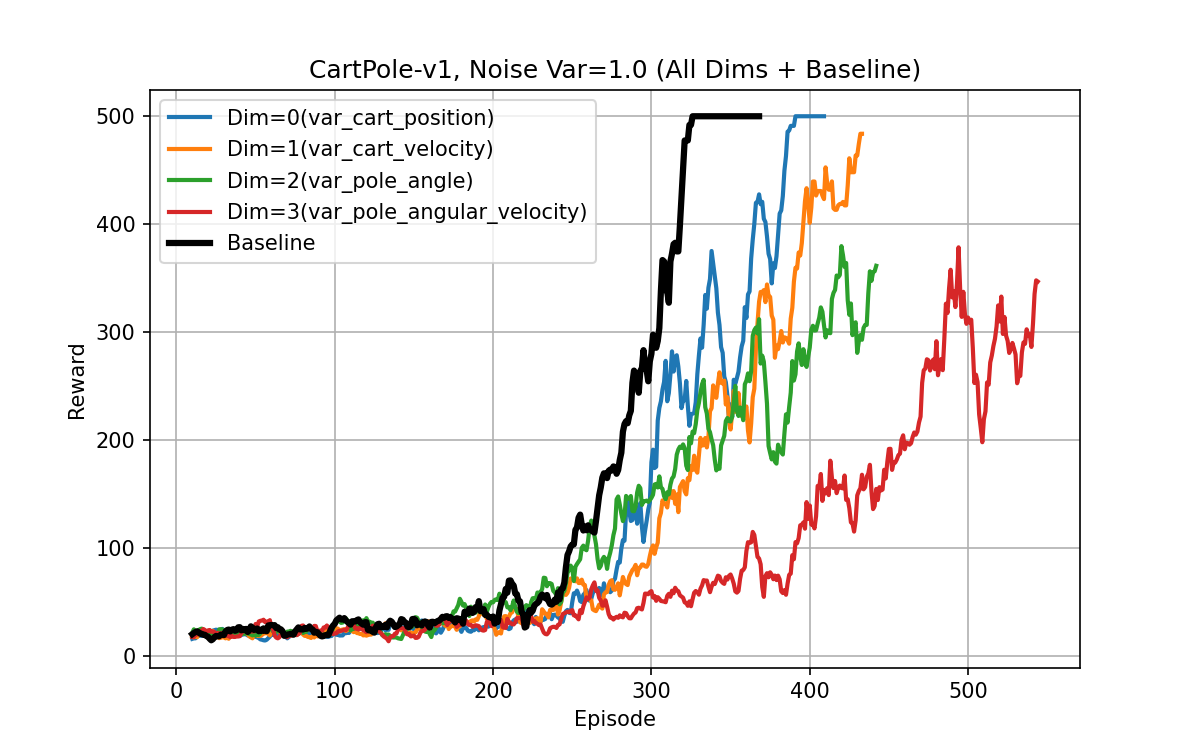}
    \caption{CartPole, $\mu=0$, $\sigma^2=1.0$}
    \label{fig:cartpole_noise_1.0}
  \end{subfigure}\\[6pt]
  \begin{subfigure}[b]{0.45\textwidth}
    \includegraphics[width=\linewidth]{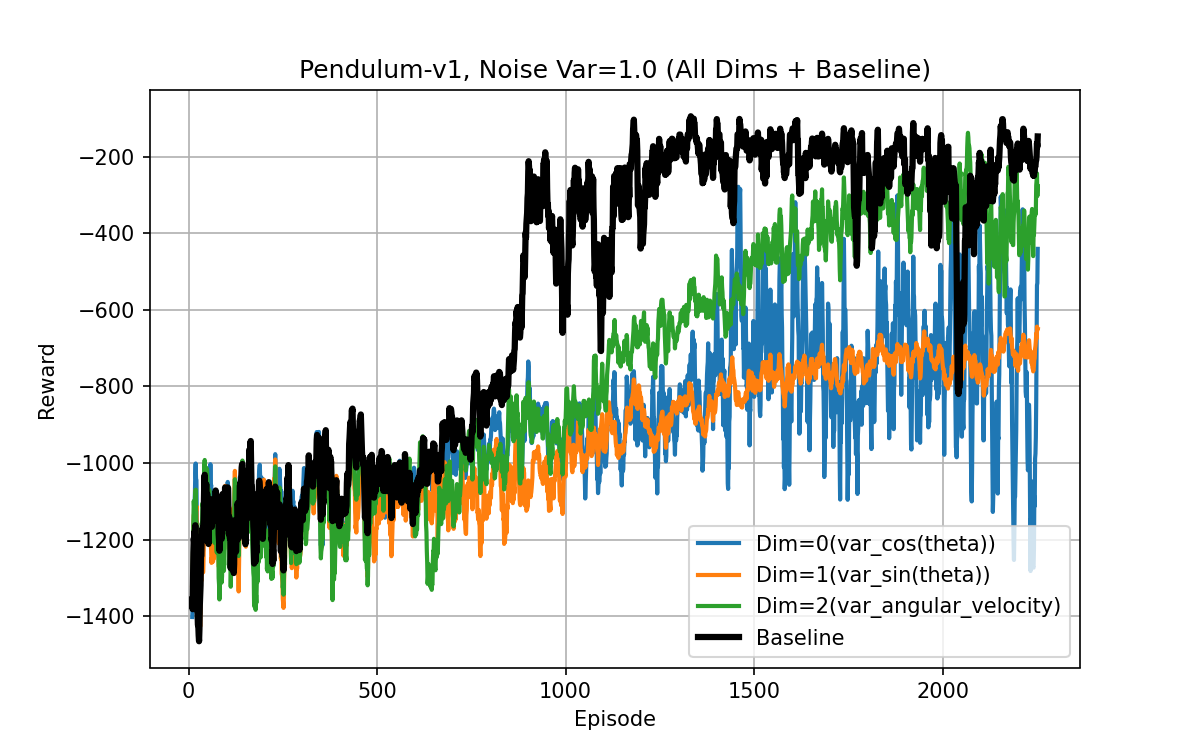}
    \caption{Pendulum, $\mu=0$, $\sigma^2=1.0$}
    \label{fig:pendulum_noise_1.0}
  \end{subfigure}
  \begin{subfigure}[b]{0.45\textwidth}
    \includegraphics[width=\linewidth]{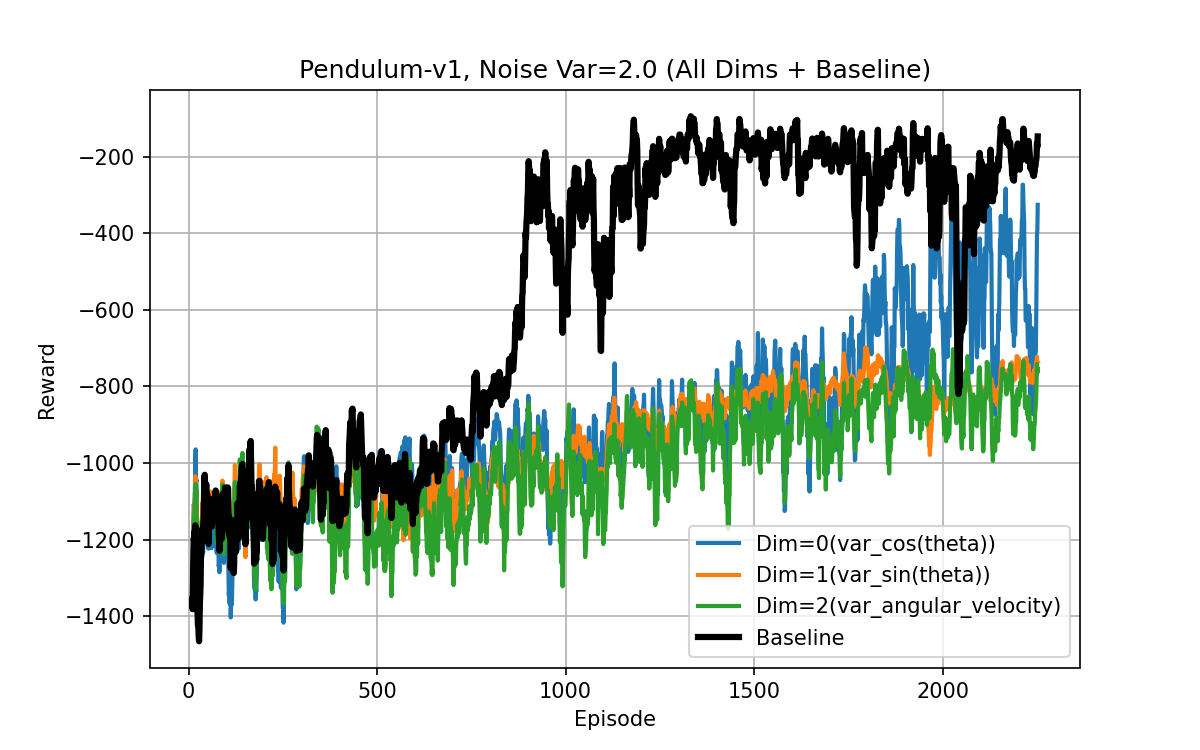}
    \caption{Pendulum, $\mu=0$, $\sigma^2=2.0$}
    \label{fig:pendulum_noise_2.0}
  \end{subfigure}\\[6pt]
  \begin{subfigure}[b]{0.45\textwidth}
    \includegraphics[width=\linewidth]{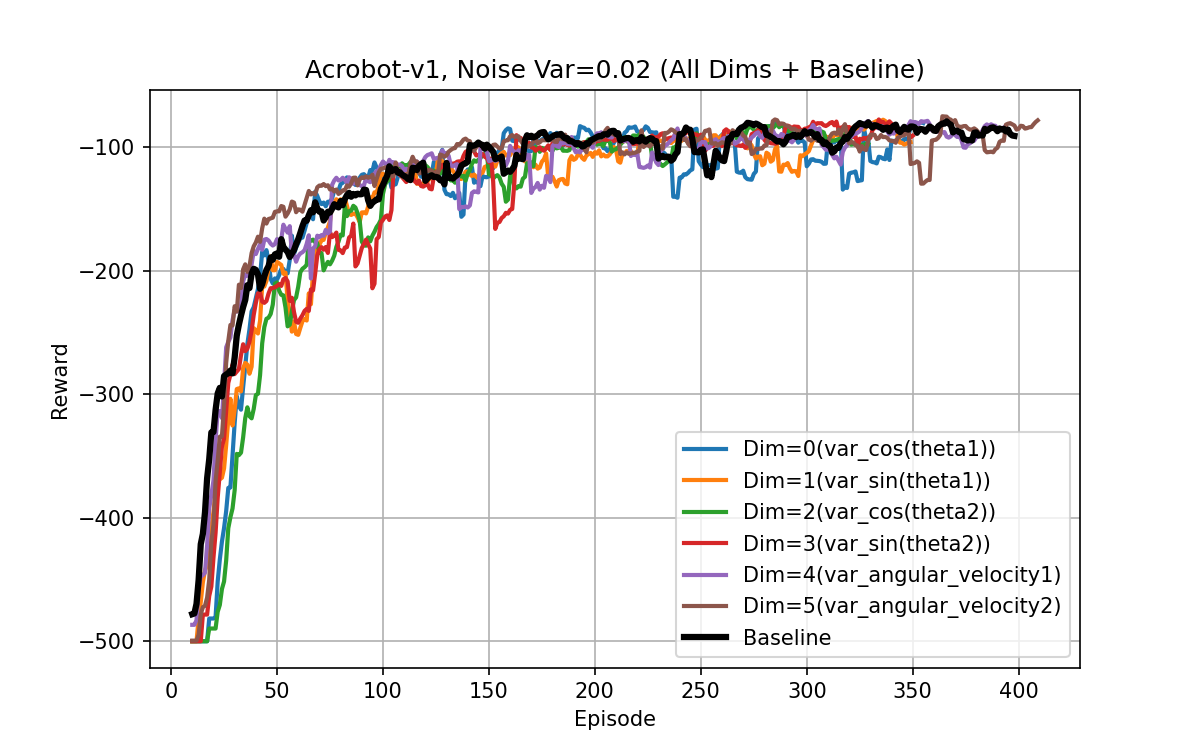}
    \caption{Acrobot, $\mu=0$, $\sigma^2=0.02$}
    \label{fig:acrobot_noise_0.02}
  \end{subfigure}
  \begin{subfigure}[b]{0.45\textwidth}
    \includegraphics[width=\linewidth]{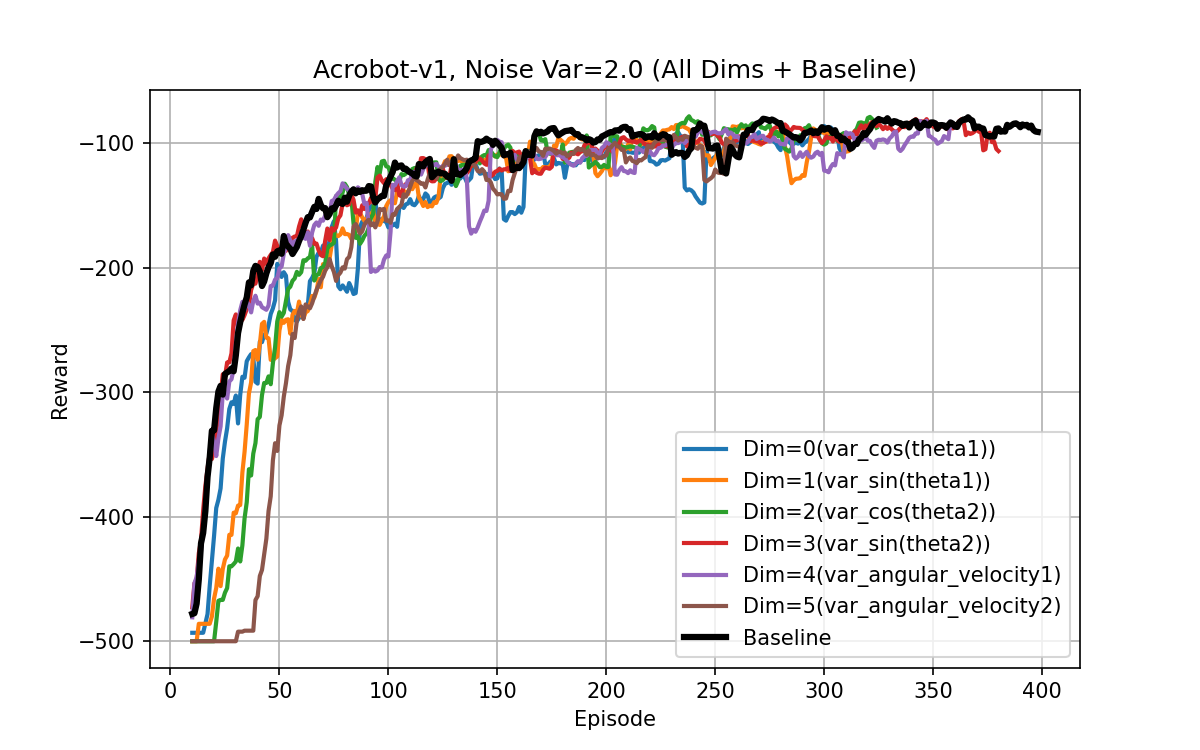}
    \caption{Acrobot, $\mu=0$, $\sigma^2=2.0$}
    \label{fig:acrobot_noise_2.0}
  \end{subfigure}
  \caption{\textbf{Effects of i.i.d.\ Gaussian Noise.} Each panel compares the noise‐free \textit{Baseline} (black) to one or more noise‐injected settings. \textbf{(a,b)}~For CartPole, a small variance $(0.02)$ barely disrupts training, but a larger variance $(1.0)$ notably impairs performance when critical dimensions (pole angle or velocity) are perturbed. \textbf{(c,d)}~Pendulum is more sensitive overall; moderate noise $(1.0)$ already degrades returns, and high noise $(2.0)$ amplifies volatility. \textbf{(e,f)}~Acrobot remains relatively robust, with minimal slowdowns even at higher noise levels. Overall, certain state dimensions (e.g., angles or angular velocities) are more vulnerable to noise, higher $\sigma^2$ typically delays learning or reduces reward, and the noise‐free Baseline continues to provide the fastest and most stable convergence.}
  \label{fig:gaussian_noise_to_observation}
\end{figure}

To evaluate how i.i.d.\ Gaussian perturbations affect both policy performance and Markov consistency, each observation dimension \(o_t^{(i)}\) may be augmented by independent draws \(\eta_t^{(i)}\sim \mathcal{N}(\mu,\sigma^2)\). The PPO agent is then trained on \(\widetilde{o}_t^{(i)} = o_t^{(i)} + \eta_t^{(i)}\) for any targeted dimension \(i\). Figure~\ref{fig:gaussian_noise_to_observation} presents the resulting learning curves for varying noise levels in CartPole, Pendulum, and Acrobot. Small noise (e.g., \(\sigma^2=0.02\)) often leaves returns near baseline levels, but larger noise (e.g., \(\sigma^2=1.0\) or \(\sigma^2=2.0\)) provokes substantial performance drops, particularly when critical angles or velocities are corrupted. Acrobot remains comparatively robust, with only minor slowdowns under higher noise.

\subsubsection{State-Space Noise Effects and MVS}\label{subsec:obs-noise}
\begin{figure}[htbp]
  \centering
  \begin{subfigure}[b]{0.45\textwidth}
    \includegraphics[width=\linewidth]{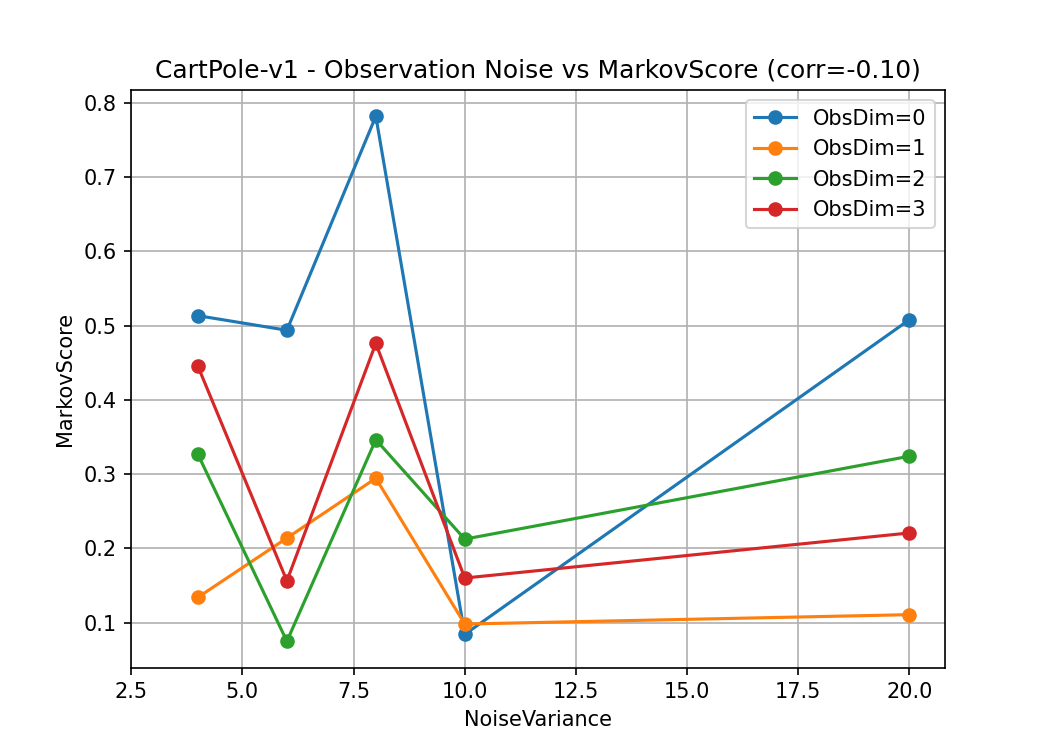}
    \caption{CartPole}
    \label{fig:cartpole_obs_noise}
  \end{subfigure}
  \begin{subfigure}[b]{0.45\textwidth}
    \includegraphics[width=\linewidth]{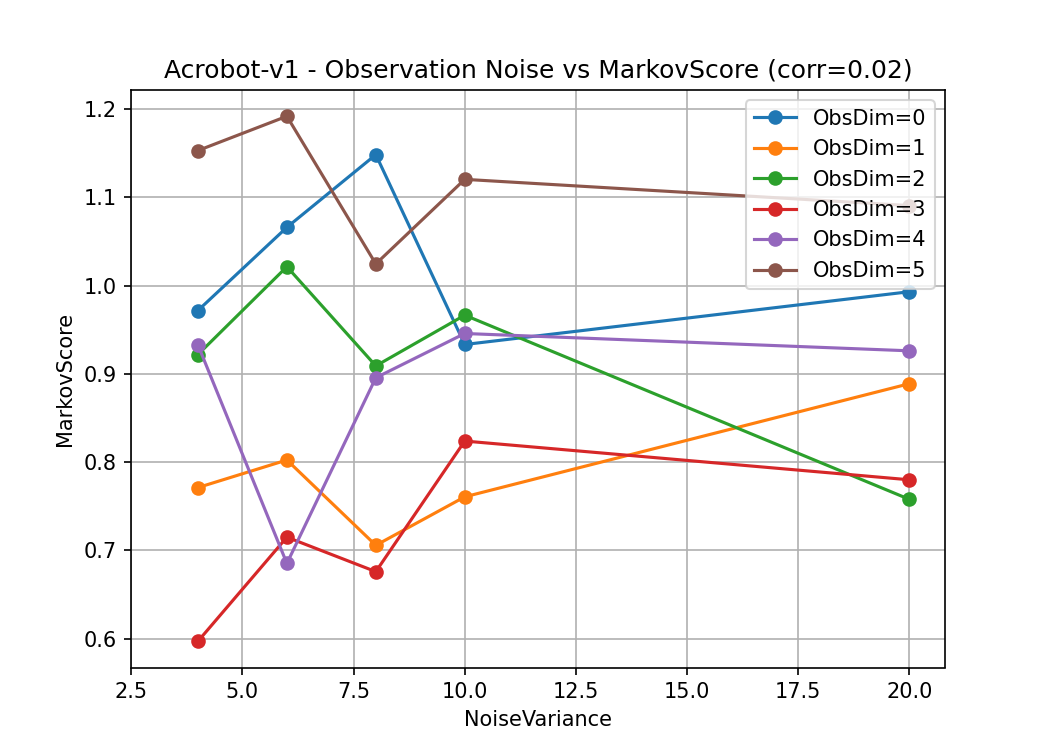}
    \caption{Acrobot}
    \label{fig:acrobot_obs_noise}
  \end{subfigure}
  \caption{\textbf{Obs Noise vs.\ MVS (i.i.d.) in CartPole and Acrobot.} Even with large variance degrading rewards, no strong multi-lag correlations are detected in either environment.}
  \label{fig:obs_noise_comparison}
\end{figure}

\begin{figure}[htbp]
  \centering
  \includegraphics[width=.45\textwidth]{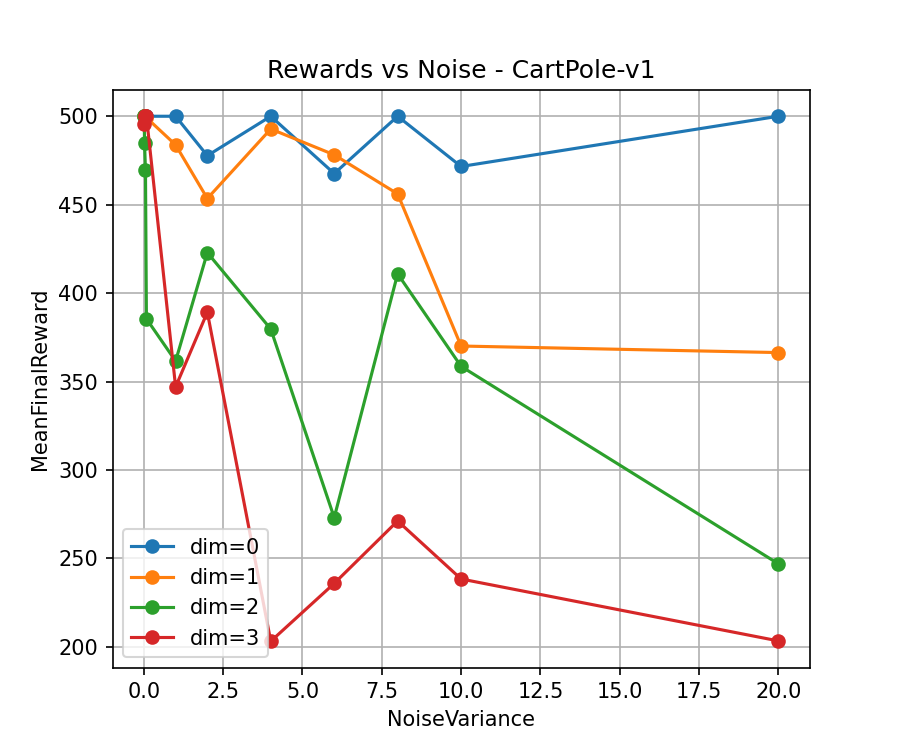}\quad
  \includegraphics[width=.45\textwidth]{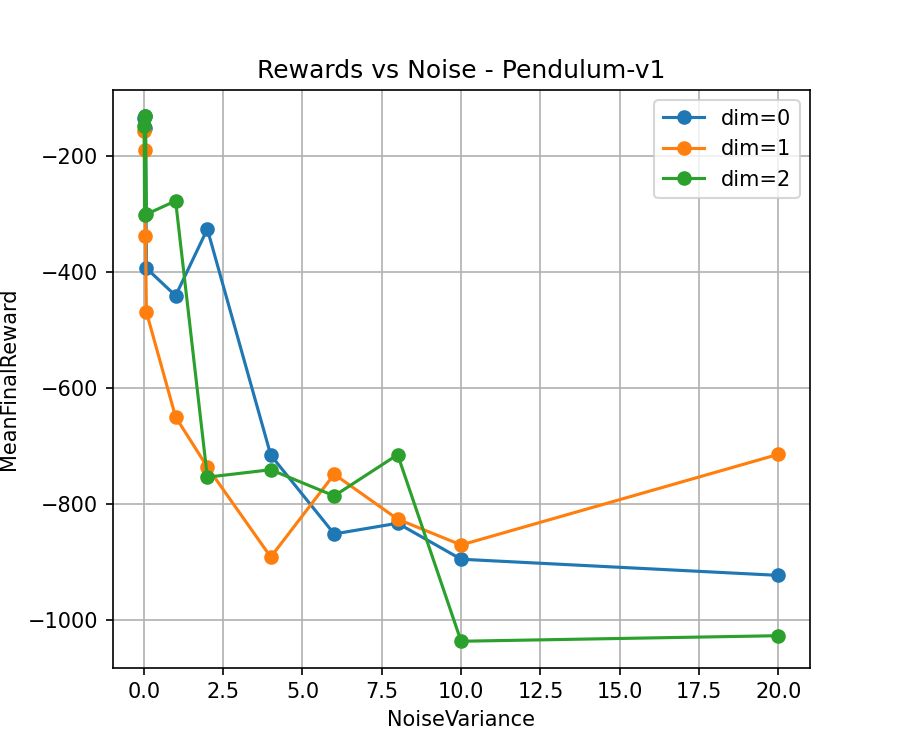}
  \caption{\textbf{Rewards vs.\ noise (i.i.d.\ Gaussian).} CartPole collapses past moderate noise, while Pendulum degrades more gradually. However, MVS often remains low despite performance drops.}
  \label{fig:rewards-vs-noise}
\end{figure}

Although elevated noise (\(\sigma^2\ge1.0\)) clearly degrades rewards (Figure~\ref{fig:rewards-vs-noise}), the Markov property remains fairly intact in i.i.d.\ Gaussian settings: PCMCI rarely uncovers strong lag‐\( \ge2 \) correlations unless the variance is extremely high. As illustrated in Figure~\ref{fig:cartpole_obs_noise}, changes in MVS remain modest for i.i.d.\ noise in CartPole, revealing that episodic returns can drop substantially even while MVS hovers near zero. These observations suggest that purely independent noise often fails to violate first‐order structure, motivating the introduction of correlated (AR) disturbances to elicit stronger multi‐lag dependencies.

\subsection{Autoregressive Noise Injection}\label{subsec:ar-noise-method}
\begin{figure}[htbp]
  \centering
  \begin{subfigure}[b]{0.45\textwidth}
    \includegraphics[width=\linewidth]{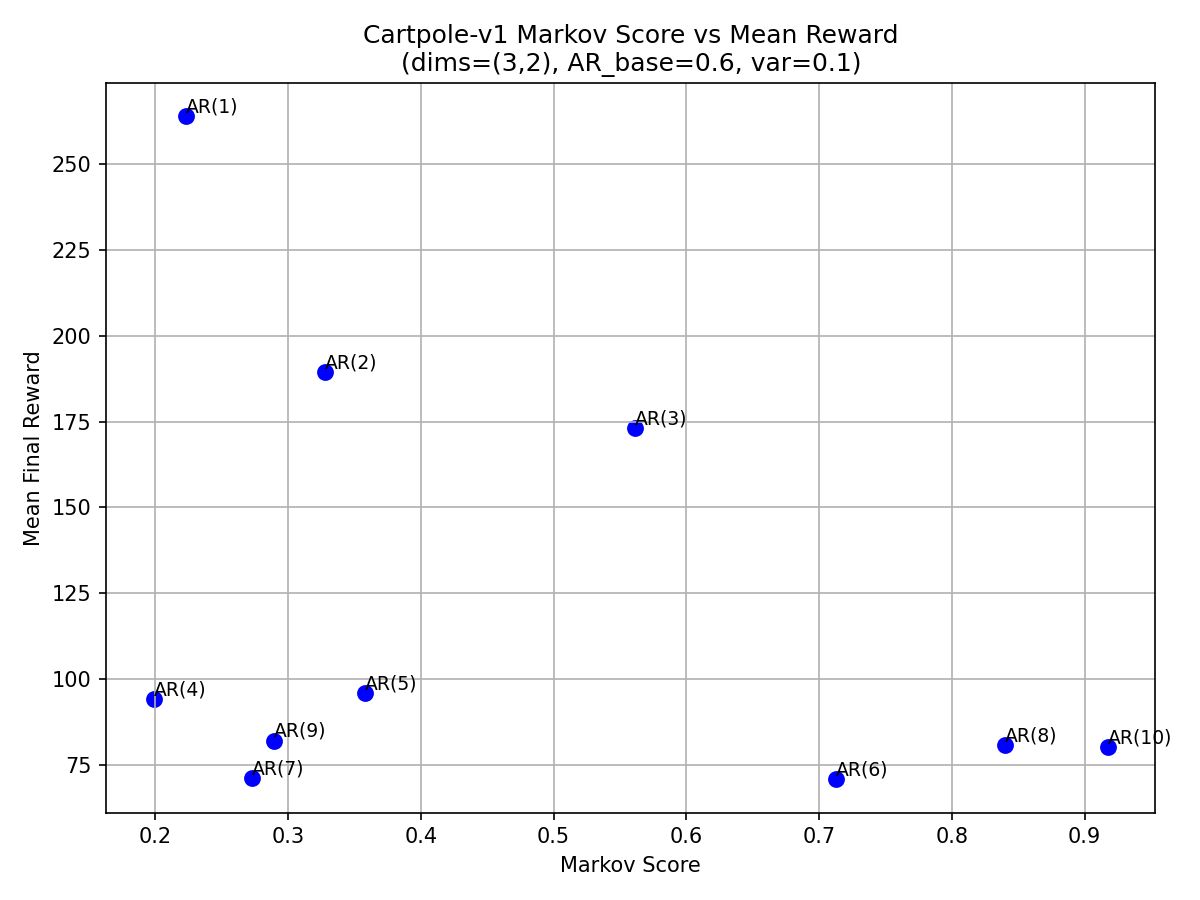}
    \caption{CartPole}
    \label{fig:cartpole_ar_order_vs_markov}
  \end{subfigure}\quad
  \begin{subfigure}[b]{0.45\textwidth}
    \includegraphics[width=\linewidth]{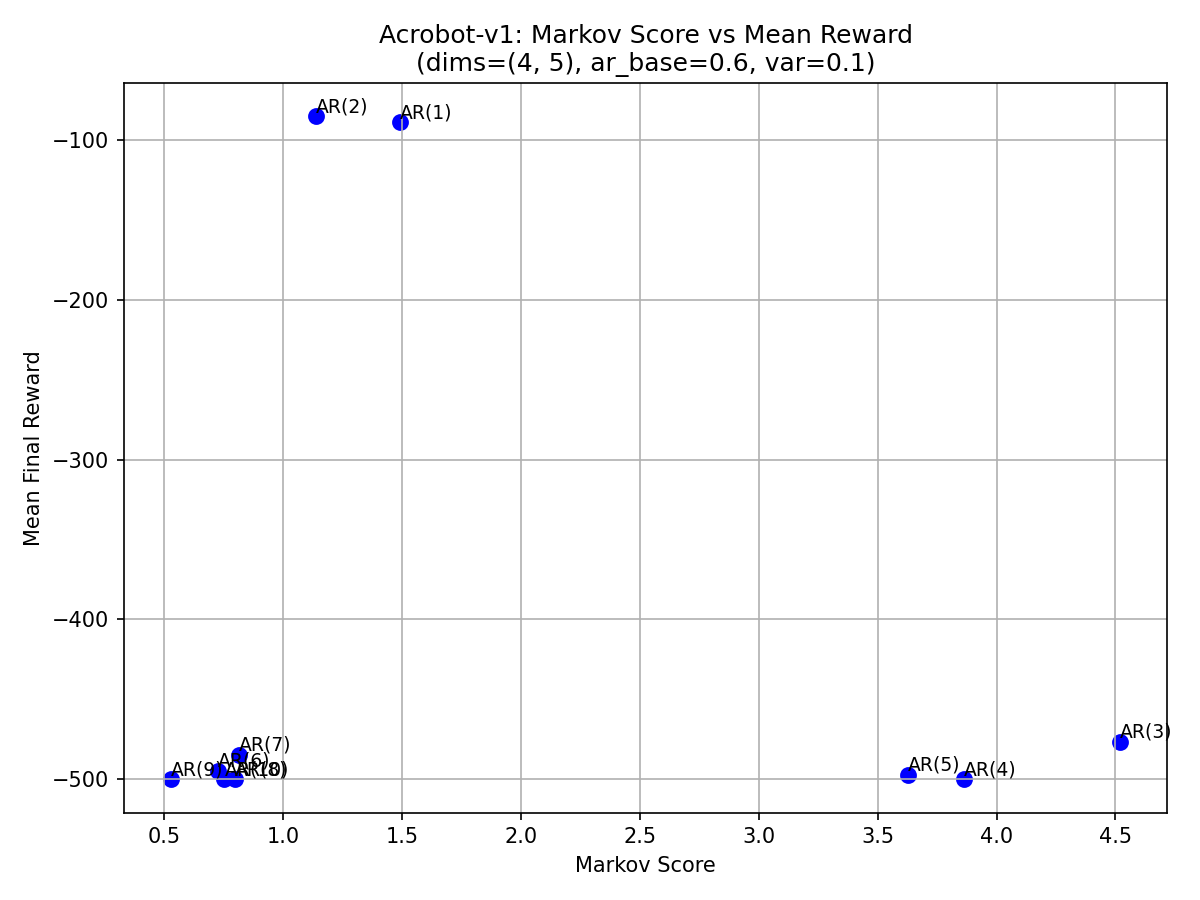}
    \caption{Acrobot}
    \label{fig:acrobot_ar_order_vs_markov}
  \end{subfigure}
  \caption{\textbf{Rewards vs.\ MVS.} Larger autoregressive orders (AR(p)) inject more multi-step correlations, driving up the Markov Violation Score (MVS) and lowering mean final rewards in both CartPole and Acrobot. In the CartPole panel, points at lower MVS (e.g., near 0.2–0.3) achieve high rewards (200–250), while points at higher MVS (0.7–0.9) sink below 100. In Acrobot, moderate MVS (0.5–0.6) yields near-optimal performance (-100), but MVS values above 3.0 correlate with scores around -400 to -500. An additional cluster forms in the lower-left corner where early termination limits PCMCI's detection window and keeps MVS from rising further. Overall, these plots confirm that as AR(p) disrupts first-order dynamics more strongly, MVS climbs and final returns drop, revealing how deviations from the Markov assumption undermine one-step RL algorithms.}
  \label{fig:ar_noise_vs_markov_reward}
\end{figure}

To induce more pronounced deviations from the first‐order Markov assumption, \emph{autoregressive} (AR) noise is introduced. Let \(\{z_t\}\) be a one-dimensional AR($p$) process,
\[
    z_{t+1} \;=\; \sum_{\ell=0}^{p-1}\rho_\ell\,z_{t-\ell} + \epsilon_t, \quad \epsilon_t \sim \mathcal{N}(0,\sigma^2).
\]
This hidden variable \(z_{t+1}\) is added to designated \emph{observation} dimensions each step, coupling consecutive states and frequently generating lag‐\( \ge2 \) dependencies. Experiments varying \(p\) and \(\rho_0\) confirm that higher AR orders and larger coefficients correlate with elevated MVS and significant performance degradation (Figure~\ref{fig:ar_noise_vs_markov_reward}). Thus, while i.i.d.\ Gaussian noise alone might not break the single‐step property, AR noise reliably induces multi‐lag correlations and accentuates the link between MVS and poorer returns.

\begin{figure}[htbp]
  \centering
  \begin{subfigure}[b]{0.45\textwidth}
    \includegraphics[width=\linewidth]{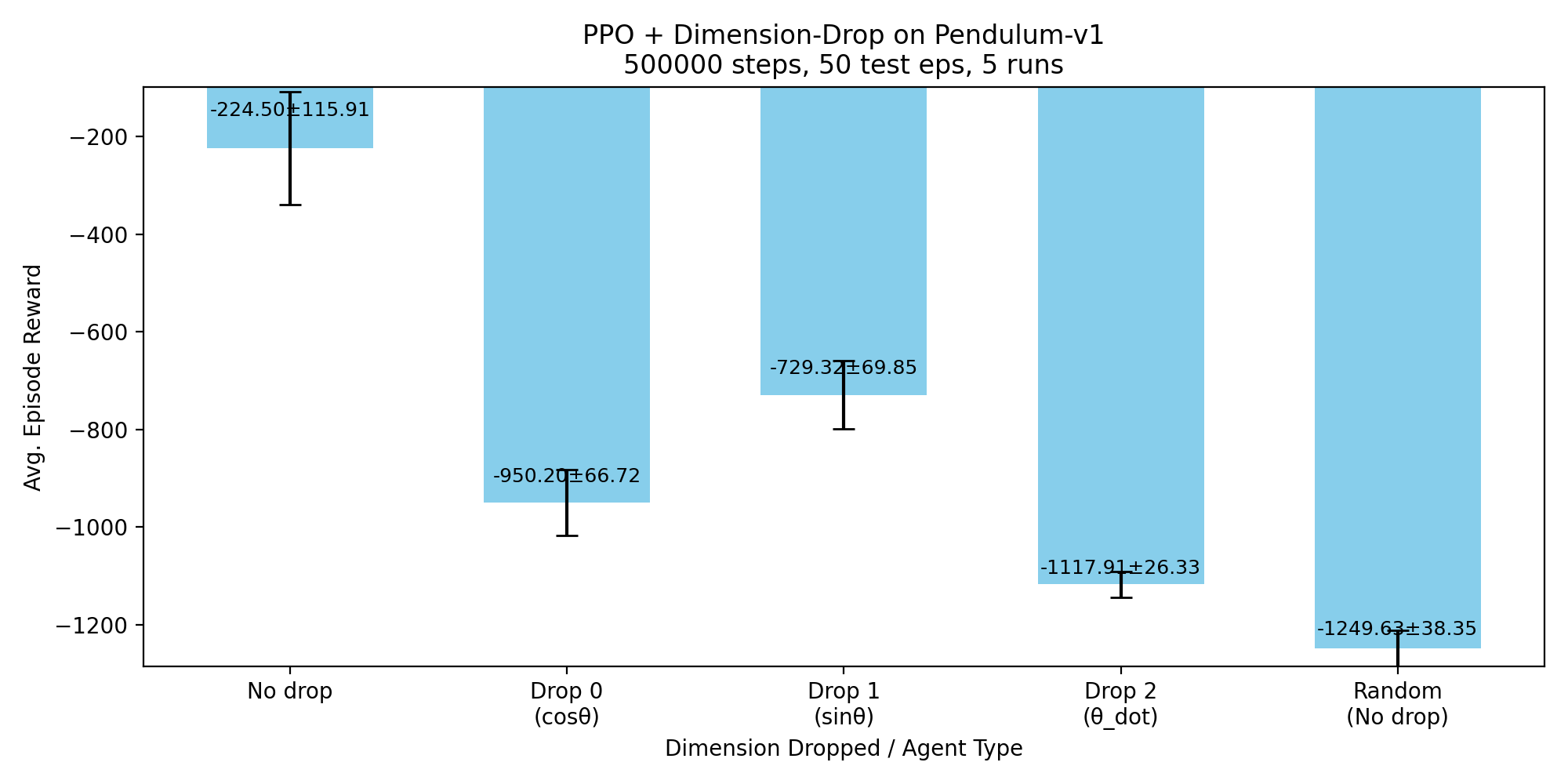}
    \caption{Pendulum returns}
    \label{fig:pendulum_drop_experiment_bar}
  \end{subfigure}\quad
  \begin{subfigure}[b]{0.45\textwidth}
    \includegraphics[width=\linewidth]{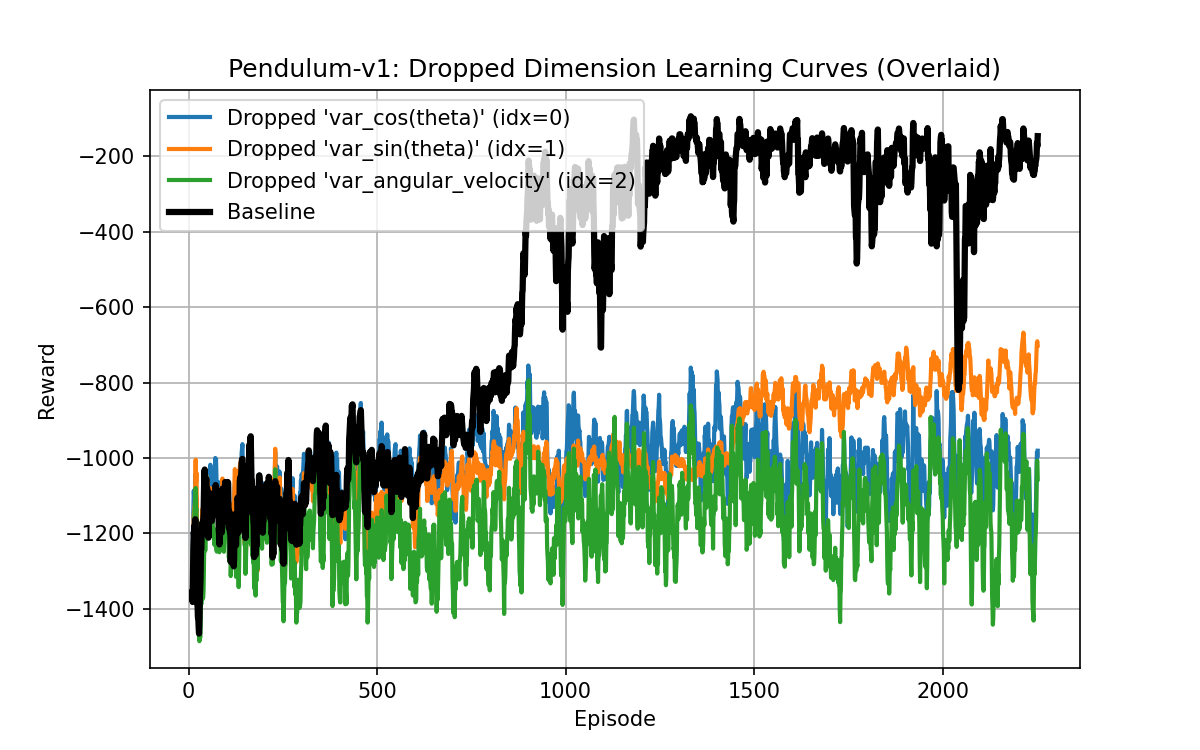}
    \caption{Pendulum learning curves}
    \label{fig:pendulum_drop_experiment_overlaid}
  \end{subfigure}
  \\[6pt]
  \begin{subfigure}[b]{0.45\textwidth}
    \includegraphics[width=\linewidth]{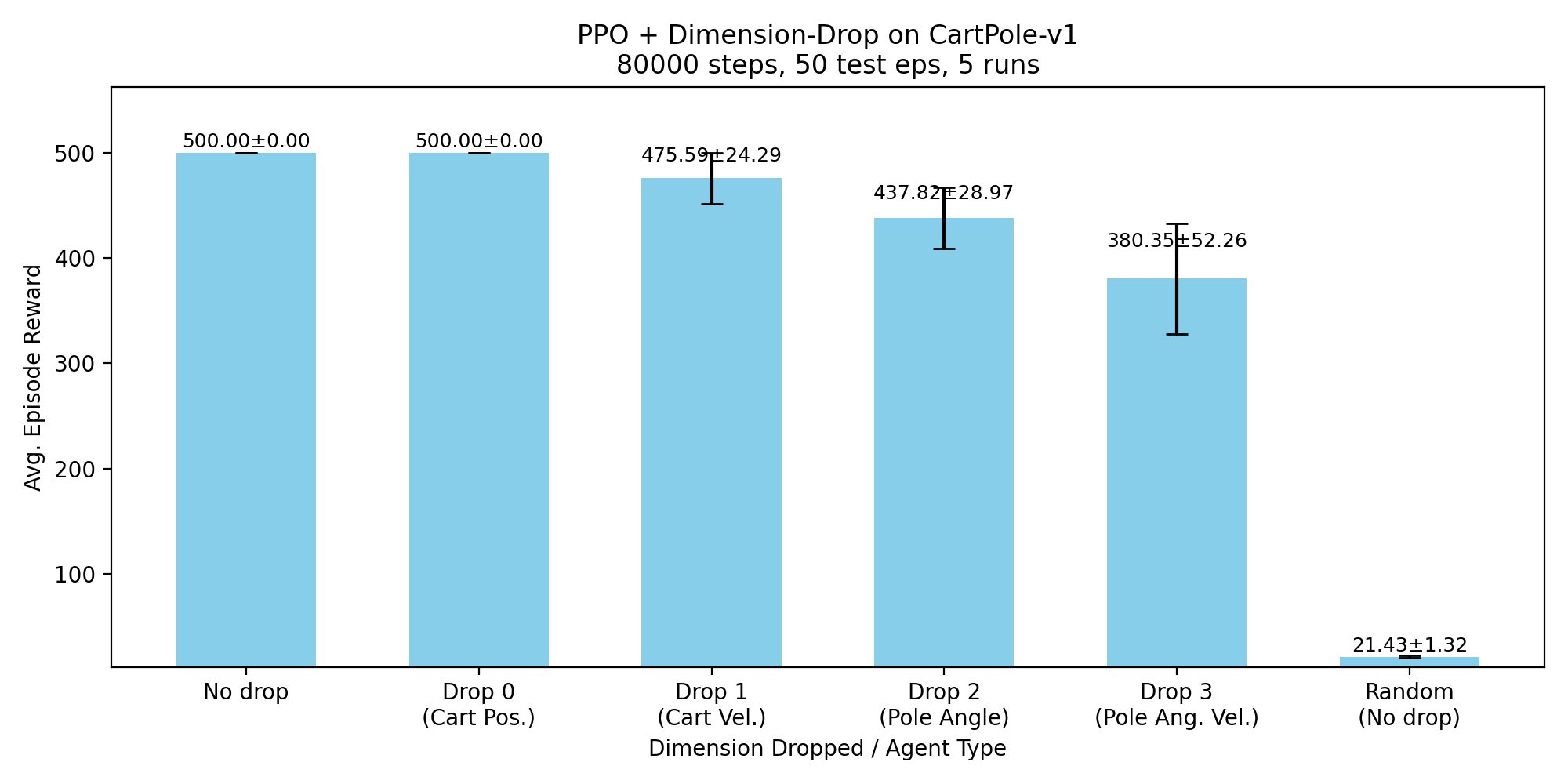}
    \caption{CartPole returns}
    \label{fig:cartpole_drop_experiment_bar}
  \end{subfigure}\quad
  \begin{subfigure}[b]{0.45\textwidth}
    \includegraphics[width=\linewidth]{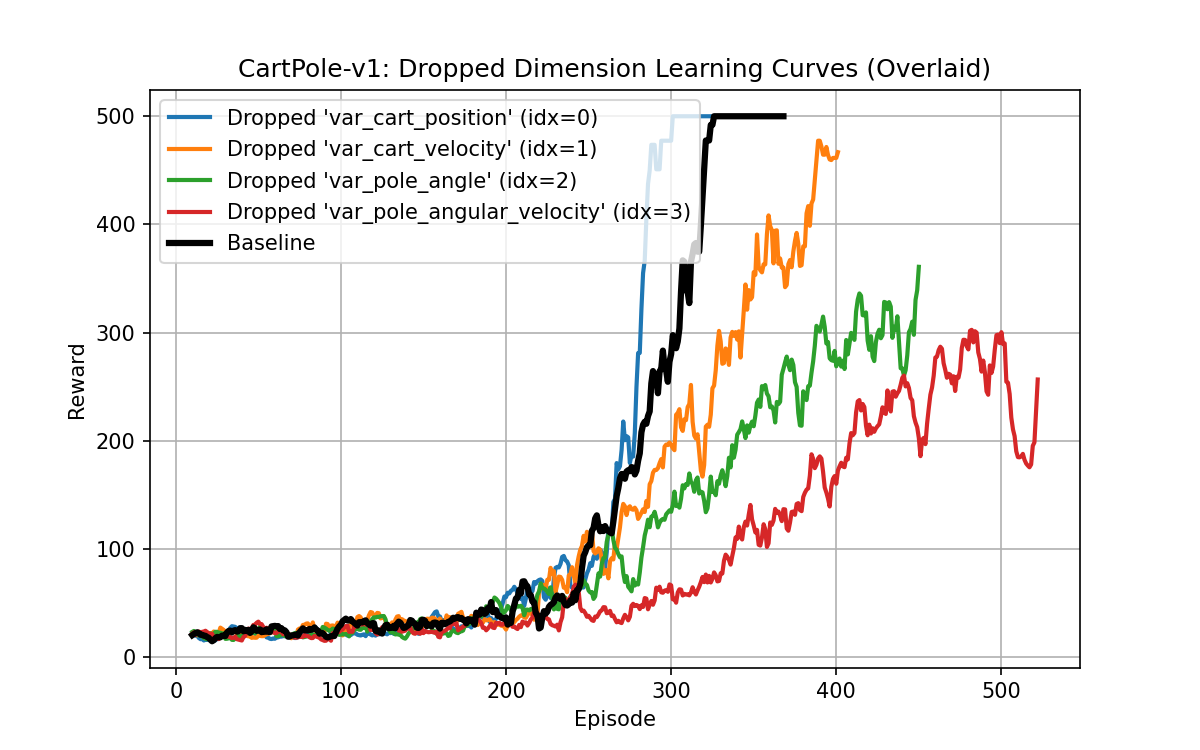}
    \caption{CartPole learning curves}
    \label{fig:cartpole_drop_experiment_overlaid}
  \end{subfigure}
  \\[6pt]
  \begin{subfigure}[b]{0.45\textwidth}
    \includegraphics[width=\linewidth]{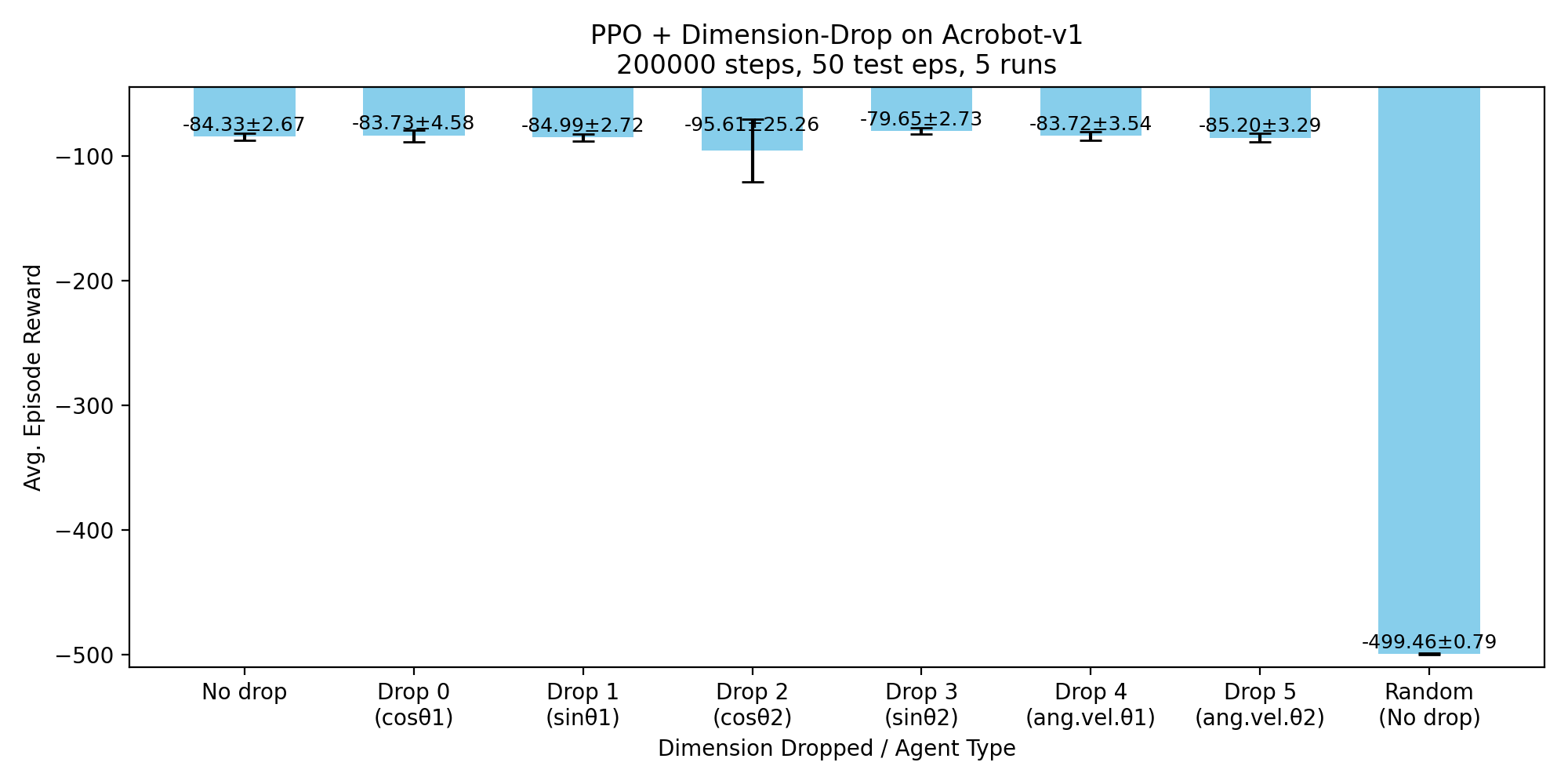}
    \caption{Acrobot returns}
    \label{fig:acrobot_drop_experiment_bar}
  \end{subfigure}\quad
  \begin{subfigure}[b]{0.45\textwidth}
    \includegraphics[width=\linewidth]{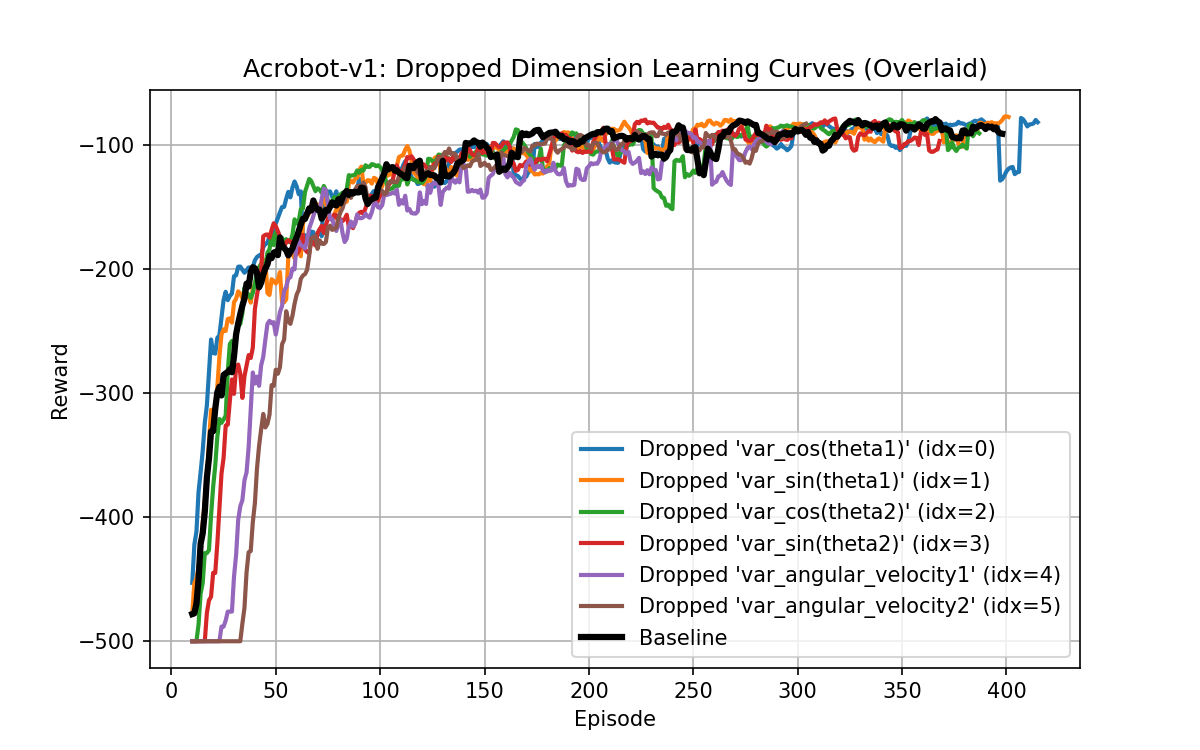}
    \caption{Acrobot learning curves}
    \label{fig:acrobot_drop_experiment_overlaid}
  \end{subfigure}
  \caption{\textbf{Dimension-Dropping Experiments.} Across Pendulum, CartPole, and Acrobot, dropping any single dimension from the observation space degrades performance compared to the Baseline, albeit to varying degrees. Pendulum’s average returns (less negative is better) become noticeably worse when dimensions such as \(\cos(\theta)\) or angular velocity are omitted, while CartPole’s returns (with a maximum of 500) fall sharply if the dropped dimension is pole angle or pole angular velocity. Acrobot, in contrast, remains comparatively robust, showing only minor changes in returns. These differences indicate that each environment depends uniquely on certain state variables for effective control, with Pendulum and CartPole hinging on angular and velocity information. Overall, removing crucial dimensions can significantly impair learning, highlighting the importance of these variables. Sensitivity also varies by environment, as Pendulum and CartPole exhibit steeper performance drops, whereas Acrobot tolerates dimension loss relatively well.}
  \label{fig:dim-drop-results}
\end{figure}

\subsection{Dimension Dropping}\label{subsec:dimension-dropping}
Motivated by the i.i.d.\ Gaussian noise results (Figure~\ref{fig:gaussian_noise_to_observation}), which hinted that certain dimensions were more critical, the next step was to drop each dimension entirely and observe any performance shifts. This is akin to senor malfunction or total loss of signal in real world systems. Surprisingly, removing some dimensions (e.g., cart position in CartPole or various joint components in Acrobot) produced negligible changes, revealing a degree of redundancy in those tasks. In contrast, omitting more pivotal features—such as pole angle or angular velocity in CartPole, or the angular velocity in Pendulum—triggered substantial performance drops and elevated Markov Violation Scores. These findings confirm that although certain state variables can be safely excluded, others are indispensable for first‐order RL methods to operate effectively.

\subsection{Overall Analysis: Correlating MVS and Policy Performance}\label{subsec:mvs-vs-performance}
Collectively, these experiments show how MVS correlates with (and often predicts) policy breakdown. In \emph{CartPole} and \emph{Pendulum}, large perturbations to crucial dimensions (e.g., pole angles or angular velocities) often raise MVS and reduce returns drastically; by contrast, \emph{Acrobot} exhibits greater redundancy, tolerating moderate distortions or dropped variables without catastrophic failure. Monitoring MVS alongside standard reward curves thus flags emergent multi‐lag dependencies in non-Markov settings (e.g., under AR noise or dimension-critical omissions). Such insights can guide robust controller design and inform representation learning, ensuring that the most causally pivotal features remain intact for stable, first-order RL.
\section{Limitations and Future Work}
\label{sec:limitations-future}

While these experiments demonstrate ways in which noise and dimension changes can compromise the Markov property in popular control benchmarks, several constraints persist. Only three domains (CartPole, Pendulum, Acrobot) were examined, which restricts generalization to more advanced tasks. Concentrating solely on PPO also leaves open the question of whether other algorithms (e.g., SAC \citep{haarnoja_soft_2018}, Q-learning \citep{mnih_playing_2013}, or model-based approaches) exhibit comparable sensitivities. Both the noise and dimension modifications were relatively basic, and the Markov Violation Score (MVS) hinges on linear partial-correlation tests; extending it with more flexible metrics could improve the detection of nonlinear effects. In addition, real-world sensor faults or actuator delays were not considered. Future studies may explore higher-dimensional tasks (e.g., multi-joint robotics) where novel forms of Markov violation can appear, investigate recurrent or Bayesian techniques for mitigating noisy signals and thereby controlling MVS spikes, and apply active dimension selection to discard extraneous variables. Model-based methods and real-world deployments could further guide the construction of robust policies under partial observability and complex noise conditions.
\section{Conclusion}\label{sec:conclusion}
This study investigated how partial observability and injected noise influence Markovian assumptions in reinforcement learning, emphasizing the detection of multi‐lag dependencies through the Markov Violation Score (MVS). Experiments with standard control tasks showed that certain state variables—such as the pole angle in CartPole or angular velocity in Pendulum—are indispensable for sustaining first‐order dynamics, while other dimensions can safely be excluded. Independent Gaussian noise frequently reduced performance but did not always generate strong lag‐$\ge2$ correlations, whereas autoregressive processes consistently produced higher MVS values and more pronounced policy breakdowns. Dimension‐dropping tests likewise indicated that some tasks (e.g., Acrobot) maintain resilience even when specific components are omitted, whereas others have a strong dependency on particular features. These findings underscore the practicality of partial‐correlation methods for identifying Markov violations and point toward adaptive or model‐based techniques as potential solutions. Broadening MVS‐based evaluations to complex, sensor‐rich environments represents a promising avenue for developing robust and generalizable RL systems.
\bibliography{main}

\begin{thebibliography}{26}
\providecommand{\natexlab}[1]{#1}
\providecommand{\url}[1]{\texttt{#1}}
\expandafter\ifx\csname urlstyle\endcsname\relax
  \providecommand{\doi}[1]{DOI: #1}\else
  \providecommand{\doi}{DOI: \begingroup \urlstyle{rm}\Url}\fi

\bibitem[wie(2012)]{wiering_partially_2012}
Partially {Observable} {Markov} {Decision} {Processes}.
\newblock In Marco Wiering and Martijn Van~Otterlo (eds.), \emph{Reinforcement {Learning}: {State}-of-the-{Art}}, volume~12 of \emph{Adaptation, {Learning}, and {Optimization}}, pp.\  387--414. Springer Berlin Heidelberg, Berlin, Heidelberg, 2012.
\newblock ISBN 978-3-642-27644-6 978-3-642-27645-3.
\newblock \doi{10.1007/978-3-642-27645-3}.
\newblock URL \url{https://link.springer.com/chapter/10.1007/978-3-642-27645-3_12}.

\bibitem[Haarnoja et~al.(2018)Haarnoja, Zhou, Abbeel, and Levine]{haarnoja_soft_2018}
Tuomas Haarnoja, Aurick Zhou, Pieter Abbeel, and Sergey Levine.
\newblock Soft {Actor}-{Critic}: {Off}-{Policy} {Maximum} {Entropy} {Deep} {Reinforcement} {Learning} with a {Stochastic} {Actor}, August 2018.
\newblock URL \url{http://arxiv.org/abs/1801.01290}.
\newblock arXiv:1801.01290 [cs].

\bibitem[Hollenstein et~al.()Hollenstein, Auddy, Saveriano, Renaudo, and Piater]{hollenstein_action_nodate}
Jakob Hollenstein, Sayantan Auddy, Matteo Saveriano, Erwan Renaudo, and Justus Piater.
\newblock Action {Noise} in {Off}-{Policy} {Deep} {Reinforcement} {Learning}: {Impact} on {Exploration} and {Performance}.

\bibitem[Hollenstein et~al.(2024)Hollenstein, Martius, and Piater]{hollenstein_colored_2024}
Jakob Hollenstein, Georg Martius, and Justus Piater.
\newblock Colored {Noise} in {PPO}: {Improved} {Exploration} and {Performance} through {Correlated} {Action} {Sampling}.
\newblock \emph{Proceedings of the AAAI Conference on Artificial Intelligence}, 38\penalty0 (11):\penalty0 12466--12472, March 2024.
\newblock ISSN 2374-3468, 2159-5399.
\newblock \doi{10.1609/aaai.v38i11.29139}.
\newblock URL \url{http://arxiv.org/abs/2312.11091}.
\newblock arXiv:2312.11091 [cs].

\bibitem[Igl et~al.(2019)Igl, Ciosek, Li, Tschiatschek, Zhang, Devlin, and Hofmann]{igl_generalization_2019}
Maximilian Igl, Kamil Ciosek, Yingzhen Li, Sebastian Tschiatschek, Cheng Zhang, Sam Devlin, and Katja Hofmann.
\newblock Generalization in {Reinforcement} {Learning} with {Selective} {Noise} {Injection} and {Information} {Bottleneck}.
\newblock In \emph{Advances in {Neural} {Information} {Processing} {Systems}}, volume~32. Curran Associates, Inc., 2019.
\newblock URL \url{https://proceedings.neurips.cc/paper_files/paper/2019/hash/e2ccf95a7f2e1878fcafc8376649b6e8-Abstract.html}.

\bibitem[Laskin et~al.(2020)Laskin, Lee, Stooke, Pinto, Abbeel, and Srinivas]{laskin_reinforcement_2020}
Misha Laskin, Kimin Lee, Adam Stooke, Lerrel Pinto, Pieter Abbeel, and Aravind Srinivas.
\newblock Reinforcement {Learning} with {Augmented} {Data}.
\newblock In \emph{Advances in {Neural} {Information} {Processing} {Systems}}, volume~33, pp.\  19884--19895. Curran Associates, Inc., 2020.
\newblock URL \url{https://proceedings.neurips.cc/paper/2020/hash/e615c82aba461681ade82da2da38004a-Abstract.html}.

\bibitem[Lauri et~al.(2023)Lauri, Hsu, and Pajarinen]{lauri_partially_2023}
Mikko Lauri, David Hsu, and Joni Pajarinen.
\newblock Partially {Observable} {Markov} {Decision} {Processes} in {Robotics}: {A} {Survey}.
\newblock \emph{IEEE Transactions on Robotics}, 39\penalty0 (1):\penalty0 21--40, February 2023.
\newblock ISSN 1552-3098, 1941-0468.
\newblock \doi{10.1109/TRO.2022.3200138}.
\newblock URL \url{https://ieeexplore.ieee.org/document/9899480/}.

\bibitem[Li et~al.(2021)Li, Gupta, Reddy, Pong, Zhou, Yu, and Levine]{li_mural_2021}
Kevin Li, Abhishek Gupta, Ashwin Reddy, Vitchyr Pong, Aurick Zhou, Justin Yu, and Sergey Levine.
\newblock {MURAL}: {Meta}-{Learning} {Uncertainty}-{Aware} {Rewards} for {Outcome}-{Driven} {Reinforcement} {Learning}, July 2021.
\newblock URL \url{http://arxiv.org/abs/2107.07184}.
\newblock arXiv:2107.07184 [cs].

\bibitem[Liu et~al.(2022{\natexlab{a}})Liu, Chung, Szepesvari, and Jin]{liu_when_2022}
Qinghua Liu, Alan Chung, Csaba Szepesvari, and Chi Jin.
\newblock When {Is} {Partially} {Observable} {Reinforcement} {Learning} {Not} {Scary}?
\newblock In \emph{Proceedings of {Thirty} {Fifth} {Conference} on {Learning} {Theory}}, pp.\  5175--5220. PMLR, June 2022{\natexlab{a}}.
\newblock URL \url{https://proceedings.mlr.press/v178/liu22f.html}.
\newblock ISSN: 2640-3498.

\bibitem[Liu et~al.(2022{\natexlab{b}})Liu, Bai, Blanchet, Dong, Xu, Zhou, and Zhou]{liu_distributionally_2022}
Zijian Liu, Qinxun Bai, Jose Blanchet, Perry Dong, Wei Xu, Zhengqing Zhou, and Zhengyuan Zhou.
\newblock Distributionally {Robust} \${Q}\$-{Learning}.
\newblock In \emph{Proceedings of the 39th {International} {Conference} on {Machine} {Learning}}, pp.\  13623--13643. PMLR, June 2022{\natexlab{b}}.
\newblock URL \url{https://proceedings.mlr.press/v162/liu22a.html}.
\newblock ISSN: 2640-3498.

\bibitem[Mnih et~al.(2013)Mnih, Kavukcuoglu, Silver, Graves, Antonoglou, Wierstra, and Riedmiller]{mnih_playing_2013}
Volodymyr Mnih, Koray Kavukcuoglu, David Silver, Alex Graves, Ioannis Antonoglou, Daan Wierstra, and Martin Riedmiller.
\newblock Playing {Atari} with {Deep} {Reinforcement} {Learning}, December 2013.
\newblock URL \url{http://arxiv.org/abs/1312.5602}.
\newblock arXiv:1312.5602 [cs].

\bibitem[Ota et~al.(2020)Ota, Oiki, Jha, Mariyama, and Nikovski]{ota_can_2020}
Kei Ota, Tomoaki Oiki, Devesh Jha, Toshisada Mariyama, and Daniel Nikovski.
\newblock Can {Increasing} {Input} {Dimensionality} {Improve} {Deep} {Reinforcement} {Learning}?
\newblock In \emph{Proceedings of the 37th {International} {Conference} on {Machine} {Learning}}, pp.\  7424--7433. PMLR, November 2020.
\newblock URL \url{https://proceedings.mlr.press/v119/ota20a.html}.
\newblock ISSN: 2640-3498.

\bibitem[Panaganti et~al.(2022)Panaganti, Xu, Kalathil, and Ghavamzadeh]{panaganti_robust_2022}
Kishan Panaganti, Zaiyan Xu, Dileep Kalathil, and Mohammad Ghavamzadeh.
\newblock Robust {Reinforcement} {Learning} using {Offline} {Data}, October 2022.
\newblock URL \url{http://arxiv.org/abs/2208.05129}.
\newblock arXiv:2208.05129 [cs].

\bibitem[Pinto et~al.(2017)Pinto, Davidson, Sukthankar, and Gupta]{pinto_robust_2017}
Lerrel Pinto, James Davidson, Rahul Sukthankar, and Abhinav Gupta.
\newblock Robust {Adversarial} {Reinforcement} {Learning}.
\newblock In \emph{Proceedings of the 34th {International} {Conference} on {Machine} {Learning}}, pp.\  2817--2826. PMLR, July 2017.
\newblock URL \url{https://proceedings.mlr.press/v70/pinto17a.html}.
\newblock ISSN: 2640-3498.

\bibitem[Raﬃn et~al.()Raﬃn, Hill, Gleave, Kanervisto, Ernestus, and Dormann]{ran_stable-baselines3_nodate}
Antonin Raﬃn, Ashley Hill, Adam Gleave, Anssi Kanervisto, Maximilian Ernestus, and Noah Dormann.
\newblock Stable-{Baselines3}: {Reliable} {Reinforcement} {Learning} {Implementations}.

\bibitem[Runge(2022)]{runge_discovering_2022}
Jakob Runge.
\newblock Discovering contemporaneous and lagged causal relations in autocorrelated nonlinear time series datasets, January 2022.
\newblock URL \url{http://arxiv.org/abs/2003.03685}.
\newblock arXiv:2003.03685 [stat].

\bibitem[Schulman et~al.(2017)Schulman, Wolski, Dhariwal, Radford, and Klimov]{schulman_proximal_2017}
John Schulman, Filip Wolski, Prafulla Dhariwal, Alec Radford, and Oleg Klimov.
\newblock Proximal {Policy} {Optimization} {Algorithms}, August 2017.
\newblock URL \url{http://arxiv.org/abs/1707.06347}.
\newblock arXiv:1707.06347 [cs].

\bibitem[Shi et~al.(2020)Shi, Wan, Song, Lu, and Leng]{shi_does_2020}
Chengchun Shi, Runzhe Wan, Rui Song, Wenbin Lu, and Ling Leng.
\newblock Does the {Markov} {Decision} {Process} {Fit} the {Data}: {Testing} for the {Markov} {Property} in {Sequential} {Decision} {Making}.
\newblock In \emph{Proceedings of the 37th {International} {Conference} on {Machine} {Learning}}, pp.\  8807--8817. PMLR, November 2020.
\newblock URL \url{https://proceedings.mlr.press/v119/shi20c.html}.
\newblock ISSN: 2640-3498.

\bibitem[Spirtes et~al.(2001)Spirtes, Glymour, and Scheines]{spirtes_causation_2001}
Peter Spirtes, Clark Glymour, and Richard Scheines.
\newblock Causation, prediction, and search.
\newblock In \emph{Causation, prediction, and search}. MIT press, 2001.

\bibitem[Sutton \& Barto(1998)Sutton and Barto]{sutton_reinforcement_1998}
Richard~S Sutton and Andrew~G Barto.
\newblock \emph{Reinforcement {Learning}: {An} {Introduction}}.
\newblock The MIT Press, Cambridge, MA, 1998.

\bibitem[Wang et~al.(2020)Wang, Liu, and Li]{wang_reinforcement_2020}
Jingkang Wang, Yang Liu, and Bo~Li.
\newblock Reinforcement {Learning} with {Perturbed} {Rewards}.
\newblock \emph{Proceedings of the AAAI Conference on Artificial Intelligence}, 34\penalty0 (04):\penalty0 6202--6209, April 2020.
\newblock ISSN 2374-3468, 2159-5399.
\newblock \doi{10.1609/aaai.v34i04.6086}.
\newblock URL \url{https://ojs.aaai.org/index.php/AAAI/article/view/6086}.

\bibitem[Wang et~al.(2019)Wang, He, and Tan]{wang_robust_2019}
Yuhui Wang, Hao He, and Xiaoyang Tan.
\newblock Robust {Reinforcement} {Learning} in {POMDPs} with {Incomplete} and {Noisy} {Observations}, February 2019.
\newblock URL \url{http://arxiv.org/abs/1902.05795}.
\newblock arXiv:1902.05795 [cs].

\bibitem[Wisniewski et~al.(2024)Wisniewski, Chatzithanos, Guo, and Tsourdos]{wisniewski_benchmarking_2024}
Mariusz Wisniewski, Paraskevas Chatzithanos, Weisi Guo, and Antonios Tsourdos.
\newblock Benchmarking {Deep} {Reinforcement} {Learning} for {Navigation} in {Denied} {Sensor} {Environments}, October 2024.
\newblock URL \url{http://arxiv.org/abs/2410.14616}.
\newblock arXiv:2410.14616 [cs].

\bibitem[Yu et~al.()Yu, Fang, Peng, Qi, Zhou, and Shi]{yu_two-way_nodate}
Shuguang Yu, Shuxing Fang, Ruixin Peng, Zhengling Qi, Fan Zhou, and Chengchun Shi.
\newblock Two-way {Deconfounder} for {Off}-policy {Evaluation} in {Causal} {Reinforcement} {Learning}.

\bibitem[Zeng et~al.(2023)Zeng, Cai, Sun, Huang, and Hao]{zeng_survey_2023}
Yan Zeng, Ruichu Cai, Fuchun Sun, Libo Huang, and Zhifeng Hao.
\newblock A {Survey} on {Causal} {Reinforcement} {Learning}, June 2023.
\newblock URL \url{http://arxiv.org/abs/2302.05209}.
\newblock arXiv:2302.05209 [cs].

\bibitem[Zhu et~al.(2020)Zhu, Ng, and Chen]{zhu_causal_2020}
Shengyu Zhu, Ignavier Ng, and Zhitang Chen.
\newblock Causal {Discovery} with {Reinforcement} {Learning}, June 2020.
\newblock URL \url{http://arxiv.org/abs/1906.04477}.
\newblock arXiv:1906.04477 [cs].

\end{thebibliography}
\bibliographystyle{rlj}
\end{document}